\documentclass[11pt]{article}

\usepackage[final]{acl}

\usepackage{times}
\usepackage{latexsym}
\usepackage{amsthm}
\usepackage{dsfont}
\usepackage[T1]{fontenc}

\usepackage[utf8]{inputenc}
\usepackage{newunicodechar}
\newunicodechar{，}{,}

\usepackage{microtype}
\usepackage{pifont}

\usepackage{inconsolata}

\usepackage{graphicx}
\usepackage{amsmath}
\usepackage{amssymb}
\usepackage{booktabs}
\usepackage{colortbl}

\usepackage{listings}
\usepackage{makecell}
\usepackage{subcaption}
\usepackage{enumitem}
\usepackage{float}

\definecolor{lightpurple}{HTML}{E8E8FF}

\title{
Understanding and Preventing Entropy Collapse in RLVR with On-Policy Entropy Flow Optimization
}

\author{
    Huimin Xu$^1$, \quad Shuai Zhao$^1$, \quad Xiaobao Wu$^2$, \quad Anh Tuan Luu$^{1,3}$\footnotemark \\
    $^1$Nanyang Technological University, Singapore \\
    $^2$Shanghai Jiao Tong University, China, $^3$VinUniversity, Vietnam \\
    \texttt{\{huimin.xu, shuai.zhao, anhtuan.luu\}@ntu.edu.sg} \\
    \texttt{xiaobaowu@sjtu.edu.cn} \\
}

\begin{document}
\maketitle
\begin{abstract}
Reinforcement learning with verifiable rewards (RLVR) has become an effective paradigm for improving the reasoning ability of large language models. However, widely used RLVR algorithms, such as GRPO, often suffer from entropy collapse, leading to premature determinism and unstable optimization. Existing remedies, including entropy regularization and ratio-based clipping heuristics, either control entropy in a coarse-grained manner or rely on approximate on-policy training. In this paper, we revisit entropy collapse from a token-level entropy flow perspective. Our analysis reveals that entropy-decreasing tokens consistently outweigh entropy-increasing ones, resulting in a severely imbalanced entropy flow. This perspective provides a unified explanation of entropy collapse in existing RLVR algorithms and highlights the importance of balancing entropy dynamics. Motivated by this analysis, we propose On-Policy Entropy Flow Optimization (OPEFO), an adaptive entropy flow balancing mechanism that rescales entropy-increasing and entropy-decreasing updates according to their contributions to entropy change, while remaining strict on-policy. Experiments on six mathematical reasoning benchmarks demonstrate that OPEFO improves training stability and final performance~\footnote{Our code, data, and models are available at \url{https://github.com/Anna7355/Entropy-Flow}.}.

\end{abstract}

\section{Introduction}

Reinforcement Learning with Verifiable Rewards (RLVR) has emerged as an effective paradigm to advance reasoning capabilities in Large Language Models~\citep[LLMs, ] []{lambert2024tulu, jaech2024openai, guo2025deepseek, yang2025qwen3, team2025kimi}.
RLVR optimizes LLMs outputs via RL objectives guided by automated verifiable reward signals.
However, recent RLVR algorithms, such as Proximal Policy Optimization~\citep[PPO, ][]{ppo} and Group Relative Policy Optimization~\citep[GRPO, ][]{shao2024deepseekmath}, often suffer from \textit{\textbf{Entropy Collapse}}:
policy entropy drops in the early stage of training and continues to decline towards near zero.
This indicates the policy becomes prematurely deterministic, limiting policy exploration and thus hindering LLMs’ reasoning capabilities~\citep{cui2025entropy, rethinking}.

\begin{figure}[t!]
\centering
\includegraphics[width=\linewidth]{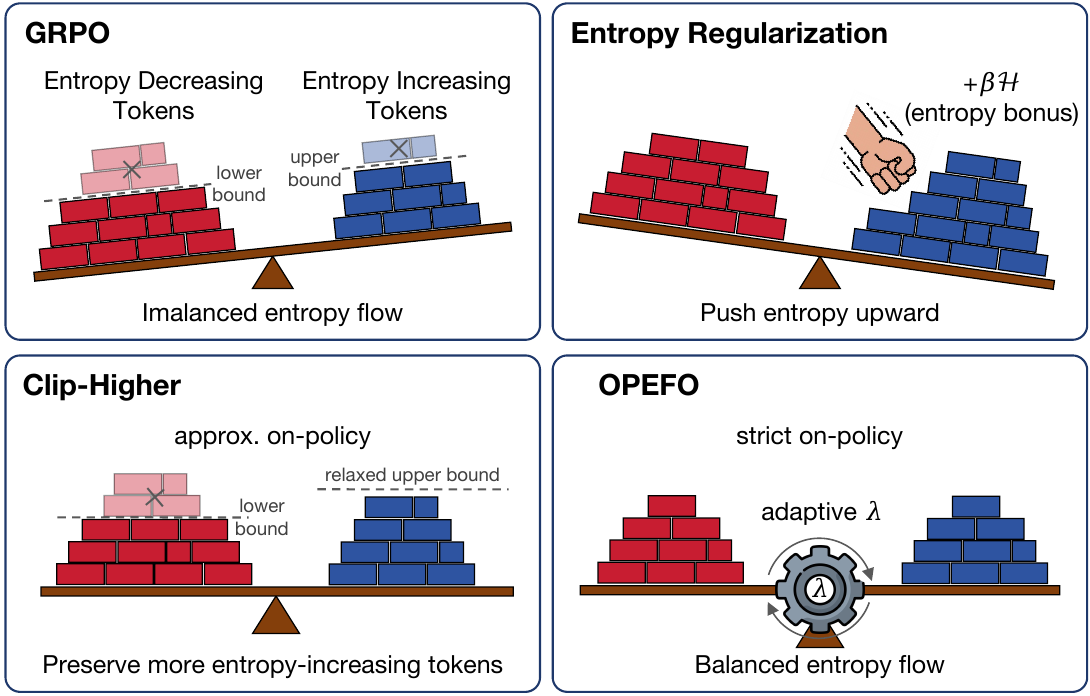}
\caption{
Entropy control mechanisms from the entropy flow perspective.
GRPO suffers from imbalanced entropy flow.
Entropy regularization increases entropy by adding an explicit entropy bonus;
Clip-higher preserves more entropy-increasing tokens via relaxing the upper clipping bound.
Differently, our OPEFO adaptively balances entropy flow via an adaptive scaling mechanism.}
\label{fig:fig1}
\end{figure}

Recent approaches to mitigating entropy collapse fall into two categories, each with notable limitations, as shown in Figure~\ref{fig:fig1}.
The first adopts entropy regularization, which introduces an explicit entropy bonus to encourage higher policy entropy~\citep{mnih2016asynchronous, haarnoja2018soft}.
But this approach indiscriminately increases entropy and may cause excessive entropy growth in later training stages~\citep{shen2025entropy, cheng2025reasoning}.
Besides, the entropy term is optimized independently of the policy-gradient objective, so it may dominate the advantage signal, leading to unstable updates and degraded performance~\citep{zhang2025maximum, liu2025prorl}.
The second category employs ratio- or clipping-based heuristics~\citep{dapo, yang2025dcpo}.
For instance, Clip-higher~\citep{dapo} relaxes the upper clipping bound of the importance sampling ratio to preserve more entropy-increasing tokens.
But they are only approximately on-policy due to their reliance on an outdated reference policy,
making them theoretically less grounded than strict on-policy optimization~\citep{baird1995residual, sutton1999policy, hao2025policy, zheng2025prosperity}.

In this paper, we address these challenges from a novel perspective.
We revisit entropy collapse through the lens of \textbf{entropy flow}: how entropy-increasing and entropy-decreasing token-level updates jointly shape overall entropy dynamics during training.
Our analysis reveals that entropy-decreasing tokens dominate the early training phase, resulting in a severely imbalanced entropy flow with a strongly negative net entropy change.
This imbalance provides a unified explanation for entropy collapse in existing RLVR algorithms.

To mitigate this imbalance issue, we propose \textbf{On-Policy Entropy Flow Optimization (OPEFO)}.
OPEFO introduces an adaptive entropy flow balancing mechanism that rescales the entropy-increasing and entropy-decreasing updates based on their contributions to entropy change.
As such, OPEFO adaptively balances the entropy flow and thus stabilizes policy entropy, as illustrated in Figure~\ref{fig:fig1}.
Moreover, it remains strict on-policy by avoiding reliance on reference policies as previous methods.
Empirical results show that OPEFO achieves best performance across two base models and six challenging mathematical reasoning benchmarks.
More importantly, it maintains the most stable entropy dynamics among strong baselines.
Overall, we summarize our contributions as follows:

\begin{itemize}[leftmargin=*]
    \item
        From an entropy flow perspective, we conduct a token-level analysis of entropy dynamics and show that entropy collapse can be interpreted as an imbalance between entropy-increasing and entropy-decreasing updates.
    \item
        We propose OPEFO, a strict on-policy entropy flow balancing mechanism that rescales entropy-increasing and entropy-decreasing token-level updates to stabilize policy entropy.
    \item
        Extensive experiments on mathematical reasoning benchmarks show that OPEFO consistently improves training stability and final performance compared to existing RLVR methods.
\end{itemize}

\section{Related Work}
Reinforcement Learning with Verifiable Rewards ~\citep[RLVR, ][]{lambert2024tulu} has recently emerged as an effective fine-tuning paradigm for large language models, achieving notable success in reasoning-intensive domains such as mathematics and programming~\citep{shao2024deepseekmath, guo2025deepseek, yang2025qwen3}. Its core idea is to replace human feedback with automatically verifiable, objective criteria as reinforcement learning rewards, thereby avoiding the cost and complexity of training human preference models. OpenAI o1~\citep{openai2024reasoning} is among the first large-scale deployments of this paradigm, demonstrating substantial performance gains on mathematical competition problems and code generation benchmarks. Following this line of work, several subsequent models—including DeepSeek-R1~\citep{guo2025deepseek}, Kimi-1.5~\citep{team2025kimi}, and Qwen-2.5~\citep{qwen2.5}—have reported matched or improved results on similar reasoning benchmarks. Overall, RLVR has shown clear advantages over prior approaches based solely on supervised fine-tuning for reasoning tasks.

These methods typically adopt GRPO-style training, yet empirical studies consistently report the emergence of entropy collapse during optimization, where the policy’s entropy rapidly diminishes as training progresses. Early approaches draw on classical entropy regularization, introducing entropy bonuses or KL penalties to stabilize optimization~\citep{mnih2016asynchronous, haarnoja2018soft}. However, recent studies show that such techniques are highly sensitive to coefficient tuning in LLM and may even mislead optimization at critical states, yielding only coarse-grained effects on entropy control~\citep{cui2025entropy, shen2025entropy}.

More recent efforts attempt to mitigate entropy collapse by modifying key components of policy optimization, including asymmetric or adaptive ratio clipping~\citep{dapo, yang2025dcpo}, selective optimization over high-entropy tokens~\citep{wang2025beyond}, or balancing positive and negative samples~\citep{zhu2025surprising}. Other methods introduce entropy-aware advantages or auxiliary objectives to encourage exploration~\citep{cheng2025reasoning, tan2025gtpo, wang2025beyond, wang2025stabilizing, deng2025decomposing}.
In this work, we adopt a token-level perspective to analyze entropy change and entropy flow, and study how these dynamics evolve under strict on-policy optimization.

\section{Preliminaries}
\label{sec:pre}

In this section, we introduce the preliminaries of RLVR and policy entropy of LLMs.

\subsection{RLVR Algorithms}
\label{sec:rlvr}
We consider reinforcement learning from verifiable rewards, where a policy
model $\pi_\theta$ autoregressively generates a token sequence
$y$ given a prompt $x$. The objective is to maximize the
expected reward received from a verifier:
\begin{equation}
\mathcal{J}(\theta)
=
\mathbb{E}_{x \sim \mathcal{D},\, y \sim \pi_\theta(\cdot \mid x)}
\big[ r(x, y) \big]
\label{eq:rlvr_obj}
\end{equation}
where $\mathcal{D}$ denotes the training data.
Following the policy gradient theorem~\citep{williams1992simple}, the gradient of
this objective can be estimated as:
\begin{equation}
\small
\nabla_\theta \mathcal{J}(\theta)
=
\mathbb{E}_{x \sim \mathcal{D},\, y \sim \pi_\theta(\cdot \mid x)}
\left[
\sum_{t=1}^{|y|}
\nabla_\theta \log \pi_\theta(y_t | x, y_{<t}) \cdot A_t
\right]
\label{eq:policy_gradient}
\end{equation}
where $A_t$ denotes the estimated advantage of token $y_t$.
To avoid training an additional value network, recent algorithms such as
GRPO~\citep{shao2024deepseekmath} estimate token-level advantages via group-wise
normalization:
\begin{equation}
A_t
=
\frac{
r(y) - \mathrm{mean}\!\left( r(y^{1:K}) \right)
}{
\mathrm{std}\!\left( r(y^{1:K}) \right)
}
\label{eq:grpo_adv}
\end{equation}
where $r(y^{1:K})$ denotes the rewards of $K$ rollouts sampled for the same prompt.
All tokens within the same response share the same normalized advantage value.

\subsection{Policy Entropy of LLMs}
For an autoregressive language model $\pi_\theta$, uncertainty at each decoding step is characterized by the entropy of the next-token distribution.
Given a state $s_t=(x, y_{<t})$, where $a$ denotes a candidate next token sampled from the vocabulary, the token-level entropy $\mathcal{H}_t$ is defined as:
\begin{equation}
\mathcal{H}_t
 = -\, \mathbb{E}_{a\sim\pi_\theta(\cdot|s_t)}
      \left[ \log \pi_\theta(a|s_t) \right]
      \label{eq:token_entropy}
\end{equation}
A smaller $\mathcal{H}_t$ indicates higher confidence, while a larger value reflects greater uncertainty or exploration.

To capture uncertainty over an entire response and dataset $\mathcal{D}$, the policy entropy is defined as the average token entropy:
\begin{equation}
\mathcal{H}(\pi_\theta, \mathcal{D})
 = \mathbb{E}_{x\sim\mathcal{D},y\sim\pi_\theta(\cdot|x)}
    \left[\frac{1}{|y|}\sum_{t=1}^{|y|} \mathcal{H}_t\right]
     \label{eq:policy_entropy}
\end{equation}
While $\mathcal{H}_t$ captures local uncertainty at individual decoding steps, $\mathcal{H}(\pi_\theta, \mathcal{D})$ provides a global summary of the policy’s exploration level.

\subsection{Token-Level Entropy Change}
\label{sec:token_entropy}
While token entropy measures the model’s uncertainty at each decoding step, entropy collapse is driven by how entropy changes during policy updates.
Following prior works~\citep{rethinking, cui2025entropy}, we therefore analyze training dynamics through token-level entropy change, decomposing the overall entropy evolution into local changes at individual decoding steps.
Directly computing the exact entropy change for large autoregressive models is intractable, due to complex dependencies across tokens. As a result, existing works commonly adopt a simplified tabular-softmax assumption, where each token’s logit is treated as conditionally independent.

\paragraph{Assumption 1} (Parameter-independent softmax).
Assume the policy $\pi_\theta$ is a tabular softmax policy, where each state-action pair $(s,a)$ is associated with an individual logit parameter $z_{s,a}(\theta)=\theta_{s,a}$.

\paragraph{Theorem 1} (First-order entropy change).
Under Assumption~1, the change
of conditional entropy between two update steps is defined as $\Delta \mathcal{H}_t \triangleq \mathcal{H}(\pi_{\theta}^{k+1}\mid s_t)-\mathcal{H}(\pi_{\theta}^{k}\mid s_t)$. Then the first-order estimation of $\Delta \mathcal{H}_t$ is:
\begin{equation}
    \begin{aligned}
        \Delta \mathcal{H}_{t} = 
        & - \eta\, \mathbb{E}_{a \sim \pi_{\theta}^{k}(\cdot \mid s_{t})} 
        \big[\, 
        A_{t} (1 - \pi_{\theta}^{k}(a \mid s_{t}))^2 \\
        & (\log \pi_{\theta}^{k}(a \mid s_{t}) + \mathcal{H}(\pi_{\theta}^{k} \mid s_{t})) \big]
    \end{aligned}
    \label{eq:delta_h}
\end{equation}
where $\eta$ is the learning rate, and $k$ indexes the policy update step.

This expression is derived from the first-order entropy change analysis introduced in~\citet{rethinking}, utilizing a Taylor expansion of the conditional entropy around the current policy logits. 
Compared to prior formulations based on importance ratios, we present the expression under the strict on-policy setting, where expectations are taken with respect to the current policy $\pi_\theta^k$ without reference policies or importance ratios.

We adopt $\Delta \mathcal{H}_t$ as a diagnostic quantity rather than a learning objective. Prior work shows that this first-order approximation closely tracks the exact entropy change in practice, justifying its use for analyzing training dynamics~\citep{rethinking}. In our setting, it enables a token-level decomposition of entropy evolution, which we leverage to identify entropy imbalance and motivate our on-policy entropy flow balancing mechanism.

\begin{figure*}[t]
    \centering
    \begin{subfigure}[t]{0.32\linewidth}
        \centering
        \includegraphics[width=\linewidth]{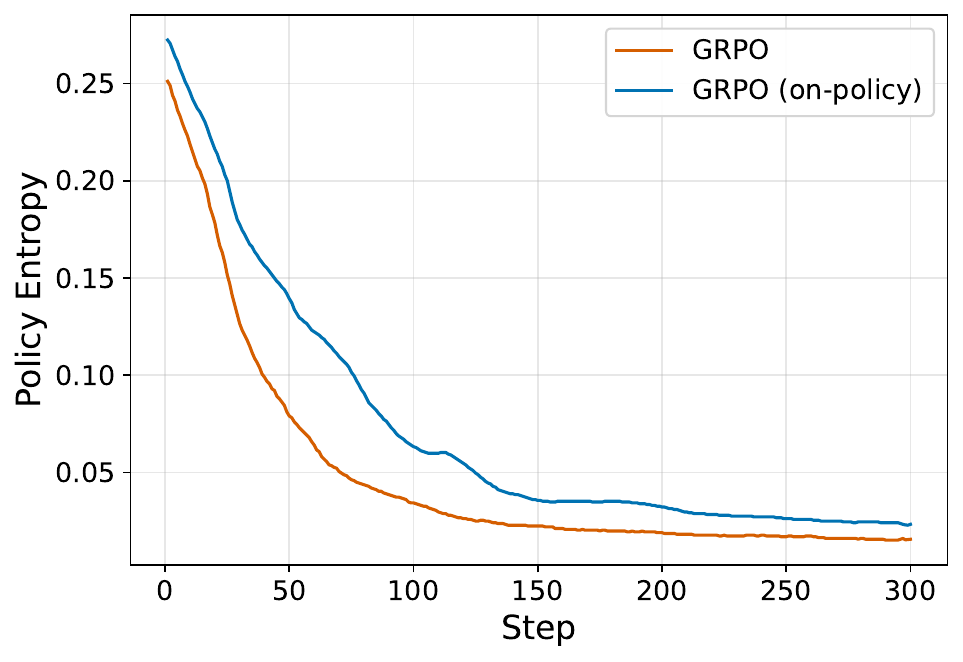}
        \caption{}
        \label{fig:grpo_entropy}
    \end{subfigure}%
    \hfill
    \begin{subfigure}[t]{0.32\linewidth}
        \centering
        \includegraphics[width=\linewidth]{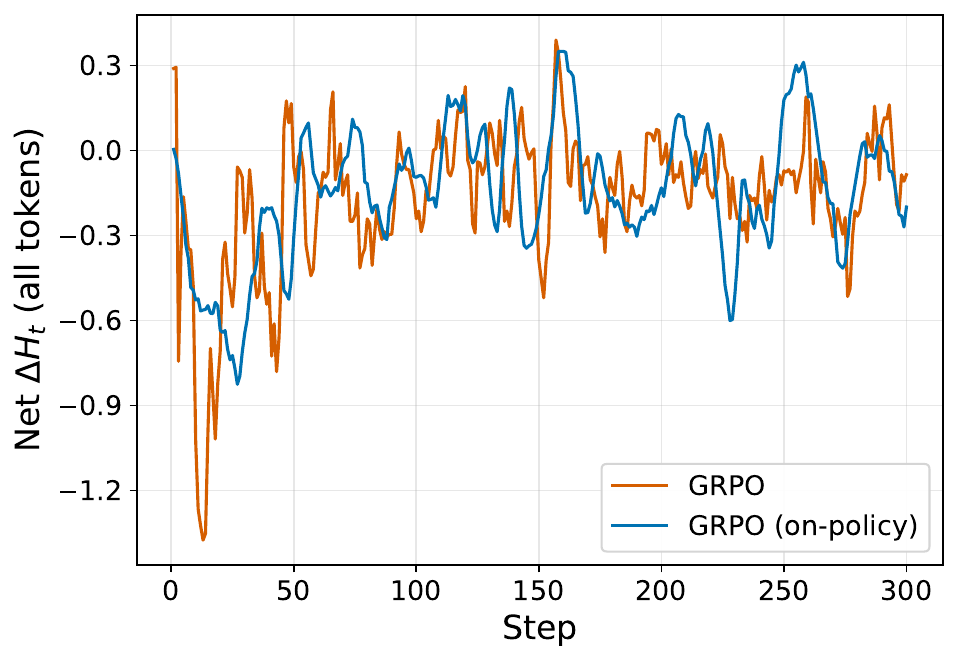}
        \caption{}
        \label{fig:grpo_delta_h}
    \end{subfigure}%
    \hfill
    \begin{subfigure}[t]{0.32\linewidth}
        \centering
        \includegraphics[width=\linewidth]{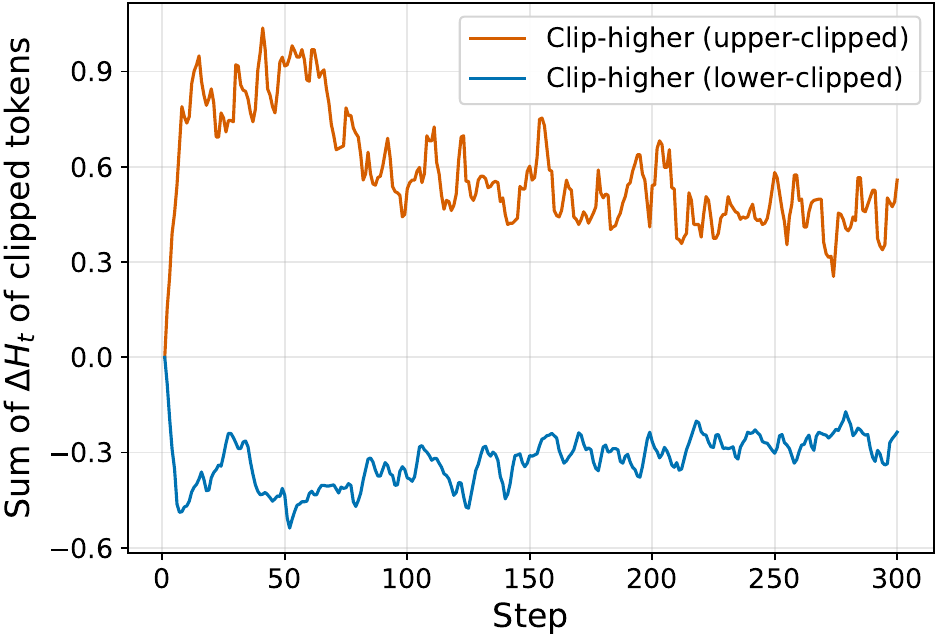}
        \caption{}
        \label{fig:clip_higher}
    \end{subfigure}
    \caption{
        Empirical analysis of entropy dynamics in GRPO and Clip-higher.
        \textbf{(a)} Policy entropy over training steps for GRPO and its variant.
        \textbf{(b)} Net $\Delta \mathcal{H}_{t}$ per update step for GRPO and its variant.
        \textbf{(c)} Sum of $\Delta \mathcal{H}_{t}$ of upper- and lower-clipped tokens under Clip-higher.
    }
    \label{fig:obs}
\end{figure*}

\section{Empirical Observations} 
\label{sec:obs}
In this section, we investigate how entropy evolves during RLVR training from an entropy flow perspective.
We examine GRPO and its strict on-policy variant.
GRPO performs 8 updates per batch, whereas the strict on-policy setting performs only one update per batch.
In addition, we analyze Clip-higher as a representative clipping-based heuristic to understand, through token-level entropy change, why such methods can alleviate entropy collapse.
All experiments are conducted using Qwen-2.5-Math-7B \cite{qwen2.5} as the base model, trained on DAPO-17k \cite{dapo}. 

\paragraph{Entropy collapse under GRPO.}
Figure~\ref{fig:obs} (a) plots the evolution of policy entropy for GRPO and its strict on-policy variant.
Both curves exhibit highly similar dynamics: starting at an entropy of approximately 0.25, dropping sharply within the first 50 training steps, and rapidly converging to nearly zero.
This observation shows that entropy collapse is a common phenomenon in GRPO training, and motivates the need for a mechanism that can stabilize entropy dynamics.

\paragraph{Token-level entropy change in GRPO.}
We compute the aggregate token-level entropy change $\sum_t \Delta \mathcal{H}_t$ at each training step, based on the Eq.~\ref{eq:delta_h}. Figure~\ref{fig:obs} (b) shows that this quantity is strongly negative during the first 50 steps, indicating that dynamics consistently bias entropy downward. This behavior directly explains the sharp entropy drop observed in Figure~\ref{fig:obs}(a). As training proceeds, the net entropy change approaches zero, coinciding with the stabilization of policy entropy. 
Together, these observations indicate that the evolution of policy entropy is driven by the cumulative token-level entropy changes, highlighting the importance of stabilizing this quantity throughout training.

\paragraph{Clipping behavior of Clip-higher.}
Since Clip-higher has demonstrated strong empirical effectiveness in alleviating entropy collapse, we analyze it as a representative clipping-based heuristic to understand its behavior from a token-level entropy flow perspective. For each training step, we compute the total $\Delta \mathcal{H}_{t}$ contributed by tokens clipped at the upper and lower bounds, respectively. 
Figure~\ref{fig:obs} (c) reveals a clear pattern: tokens clipped at the upper bound tend to exhibit positive entropy change ($\Delta \mathcal{H}_{t} > 0$), whereas those clipped at the lower bound typically exhibit negative entropy change ($\Delta \mathcal{H}_{t} < 0$).
This pattern provides a mechanistic explanation for the effectiveness of Clip-higher: by relaxing the upper bound, it selectively preserves entropy-increasing updates, thereby partially counteracting entropy collapse.
However, as this effect relies on importance-ratio clipping, it is fundamentally incompatible with strict on-policy training.

\paragraph{Summary.}
These observations suggest that entropy collapse in RLVR can be interpreted from an entropy flow perspective, where entropy-decreasing updates consistently outweigh entropy-increasing ones, leading to a strongly negative net entropy change.
Since clipping-based heuristics such as Clip-higher rely on importance-ratio clipping, they are incompatible with strict on-policy training, motivating the need for a direct mechanism to stabilize entropy dynamics under strict on-policy optimization.

\section{On-Policy Entropy Flow Optimization}
\label{sec:OPEFO}
To address the imbalanced entropy flow observed in Section~\ref{sec:obs}, we propose On-Policy Entropy Flow Optimization (OPEFO), a mechanism designed for stabilizing entropy dynamics under strict on-policy RLVR training.

\subsection{Entropy Flow Decomposition}
To formalize the entropy flow, we decompose entropy evolution into token-level contributions within each policy update. For each generated token $y_t$, we estimate the corresponding entropy change $\Delta \mathcal{H}_t$ (Eq.~\ref{eq:delta_h}) induced by a policy-gradient update. Using the sign of $\Delta \mathcal{H}_t$, we partition the token-level updates within a batch into two sets:
\begin{itemize}[leftmargin=*]
    \item \textbf{Entropy-increasing set} ($\mathcal{S}^+$): Tokens with $\Delta \mathcal{H}_t > 0$, corresponding to updates that broaden the model’s predictive distribution.
    \item \textbf{Entropy-decreasing set} ($\mathcal{S}^-$): Tokens with $\Delta \mathcal{H}_t < 0$, corresponding to updates that sharpen the distribution.
\end{itemize}
As observed in Section~\ref{sec:obs}, GRPO-style training exhibits a pronounced imbalance in entropy flow, where entropy-decreasing updates consistently outweigh entropy-increasing ones across training steps, $\sum_{t \in \mathcal{S}^-} |\Delta \mathcal{H}_t| > \sum_{t \in \mathcal{S}^+} \Delta \mathcal{H}_t$.
As this imbalance persists across updates, the policy entropy progressively collapses.

\subsection{Balanced On-Policy Objective}
To stabilize entropy dynamics under strict on-policy RLVR training, we introduce a reweighted on-policy gradient objective that rescales updates from the two token sets via a balancing coefficient $\lambda$:
\begin{equation}
\label{eq:OPEFO-obj}
\begin{aligned}
\nabla_\theta & J_{\text{OPEFO}}
(\theta)=
\mathbb{E}_{x \sim \mathcal{D},\, y \sim \pi_\theta(\cdot|x)} 
\\
&\quad
\Bigg[(1+\lambda) 
\!\!\sum_{t \in \mathcal{S}^+}
\nabla_\theta \log \pi_\theta(y_t|x,y_{<t}) \cdot A_t
\\
&\quad
+\;
(1-\lambda)
\!\!\sum_{t \in \mathcal{S}^-}
\nabla_\theta \log \pi_\theta(y_t|x,y_{<t}) \cdot A_t
\Bigg]
\end{aligned}
\end{equation}
where $\lambda \in (-1,1)$ controls the balance between entropy-increasing and entropy-decreasing updates.

This objective offers two practical advantages.
First, it is fully compatible with strict on-policy training, as it does not rely on importance sampling ratios or reference policies.
Second, by only rescaling gradient components, it preserves the original optimization direction without introducing auxiliary entropy bonuses.

\subsection{Adaptive Entropy Flow Scaling}
The remaining question is how to determine an appropriate value of $\lambda$, which we compute adaptively from batch-level entropy statistics.
Specifically, we adopt zero entropy flow as a local stabilizing criterion at the batch level. This criterion is motivated by the observation that entropy collapse arises when entropy-decreasing updates consistently outweigh entropy-increasing ones.
By enforcing a balance between these two opposing forces, we establish a condition for stabilizing entropy dynamics without external intervention. Formally, we seek a value of $\lambda$ such that:

\begin{equation}
\label{eq:entropy-balance}
\sum_{t\in \mathcal{S}^+}(1+\lambda) \Delta \mathcal{H}_{t}
+
\sum_{t\in \mathcal{S}^-}(1-\lambda)\Delta \mathcal{H}_{t}
\approx 0
\end{equation}

Solving this constraint yields a unique solution:
\begin{equation}
\label{eq:lambda-star}
\lambda^*
=
\frac{
\sum_{t\in \mathcal{S}^-}|\Delta \mathcal{H}_{t}|
-
\sum_{t\in \mathcal{S}^+}\Delta \mathcal{H}_{t}
}{
\sum_{t\in \mathcal{S}^-}|\Delta \mathcal{H}_{t}|
+
\sum_{t\in \mathcal{S}^+}\Delta \mathcal{H}_{t}
}
\end{equation}

This closed-form expression provides an adaptive entropy flow balancing coefficient computed directly from the current batch. 
We do not claim $\lambda^*$ to be globally optimal across tasks or training stages; rather, it acts as a self-adjusting mechanism that compensates for transient entropy imbalances.

\subsection{Practical Implementation}
OPEFO is straightforward to implement and requires only minimal modification to standard on-policy RLVR pipelines.
Practically, implementing OPEFO requires only a few lines of modification to standard GRPO code: compute $\Delta \mathcal{H}_{t}$, group tokens by sign, compute $\lambda^*$ using Eq.~\ref{eq:lambda-star}, and reweight gradients accordingly.
This procedure is fully compatible with strict on-policy training and introduces negligible computational overhead.
Figure~\ref{fig:opefo_code} shows the implementation code of OPEFO.

\begin{figure}[H]
\begin{lstlisting}[language=Python, basicstyle=\ttfamily\small]
def compute_opefo_loss(log_prob, advantages, delta_H, eps=1e-12, **args):
    pg_terms = - log_prob * advantages
    S_pos = delta_H > 0 
    S_neg = delta_H < 0 
    
    pos_mag = delta_H[S_pos].sum()
    neg_mag = delta_H[S_neg].abs().sum()
    
    denom = (neg_mag + pos_mag).clamp_min(eps)
    lambda_s = (neg_mag - pos_mag) / denom
    
    pg_terms[S_pos] *= (1.0 + lambda_s)
    pg_terms[S_neg] *= (1.0 - lambda_s)
    return pg_terms.mean()
\end{lstlisting}
\caption{
    Implementation code of OPEFO loss.
}
\label{fig:opefo_code}
\end{figure}

\begin{table*}[t]
\centering
\setlength{\tabcolsep}{3mm}
\resizebox{\textwidth}{!}{
\begin{tabular}{lcccccccc}
    \toprule
    \textbf{Method} & \textbf{AIME24} & \textbf{AIME25} & \textbf{AMC23} & \textbf{MATH500} & \textbf{Minerva} & \textbf{Olympiad} & \textbf{Avg.} \\
    \midrule
    \emph{Qwen2.5-Math-7B} & 13.8 & 5.3 & 44.6 & 39.6 & 9.9 & 13.8 & 21.2  \\
    GRPO & 26.7 & 13.0 & 69.8 & 80.3 & 37.1 & 46.1 & 45.5 \\
    GRPO (Strict on-policy) & 32.6 & 15.4 & 81.8 & 83.5 & 38.9 & 48.1 & 50.1 \\
    Entropy-Reg & 31.7 & 13.3 & 74.4 & 81.9 & 40.5 & 45.1 & 47.8 \\
    Clip-higher & 30.5 & 17.4 & 78.5 & 83.5 & 39.7 & 49.4 & 49.8 \\ 
    Clip–Cov & 32.2 & 18.5 & 77.9 & 85.1 & 40.3 & 50.2 & 50.7 \\
    KL–Cov& 32.6 & 17.9 & 78.3 & 84.6 & 40.9 & 48.7 &  50.5\\
    \midrule
    \textbf{OPEFO} & \textbf{34.5} & \textbf{19.2} & \textbf{82.2} & \textbf{85.3} & \textbf{41.6} & \textbf{51.8} & \textbf{52.4} \\
    \midrule
    \emph{Qwen3-Base-4B} & 9.7 & 8.8 & 51.1 & 75.4 & 33.1 & 40.5 & 36.4\\
    GRPO & 19.4 & 17.3 & 66.1 & 80.6 & 36.3 & 49.4 & 44.8\\
    GRPO (Strict on-policy) & 26.2 & 22.3 & 71.5 & 86.3 & 38.9 & 55.7 & 50.1\\
    Entropy-Reg & 22.7 & 20.8 & 70.9 & 83.4 & 37.1 & 53.3 & 48.0 \\
    Clip-higher & 23.5 & 21.7 & 70.3 & 84.8 & 39.2 & 55.3 & 49.1\\ 
    Clip–Cov & 24.2 & 23.1 & 71.9 & 85.7 & 39.7 & 56.9 & 48.9\\
    KL–Cov & 24.7 & 22.4 & 72.3 & 86.1 & 40.3 & 55.8 & 49.2\\
    \midrule
    \textbf{OPEFO} & \textbf{27.7} & \textbf{24.8} & \textbf{73.1} & \textbf{87.5} & \textbf{41.2} & \textbf{57.4} & \textbf{51.9} \\
    \bottomrule
\end{tabular}}
\caption{
Main results on mathematical reasoning benchmarks. All results are presented as percentages.
The best are in \textbf{bold}.
}
\label{tab:main_res}
\end{table*}

\section{Experiments}

\subsection{Experimental setup}
\label{sec:setup}
\paragraph{Training.}
Following prior work~\citep{rethinking, cui2025entropy}, we use Qwen2.5-Math-7B~\citep{qwen2.5} as the primary base model, and additionally include Qwen3-4B-Base~\citep{yang2025qwen3} to evaluate the generality of our method.
The training codebase is adapted from Verl~\citep{verl}. The training data is DAPO-17K~\citep{dapo}, which contains high-quality reasoning trajectories annotated with verifiable rewards.

For all methods, each rollout step samples a batch of 32 prompts with 8 responses per prompt, yielding 256 responses in total. Strict on-policy methods perform one update per rollout using the full batch, while approximate on-policy baselines follow the standard GRPO setting with 8 sequential updates over mini-batches from the same rollout data.
We optimize using AdamW~\citep{loshchilov2017decoupled}, with learning rates of 2.83e-6 for strict on-policy methods~\footnote{Following common practice, the learning rate is scaled proportionally to the square root of the effective batch size increase (i.e., $\sqrt{8}$) to maintain a comparable optimization noise scale \citep{2017Accurate, smith2017bayesian}.} and 1e-6 for approximate on-policy baselines. All models are trained with a linear warm-up over the first 10 rollout steps.
The maximum response length is set to 3000 tokens for Qwen2.5-Math-7B-base and 8000 tokens for Qwen3-4B-base. All experiments are conducted on 32 A100 GPUs.

\paragraph{Baselines.}
We compare OPEFO with several representative baselines. GRPO is used as the primary reference method, with both the upper and lower clipping bounds set to 0.2. Entropy regularization (Entropy-Reg) augments the objective with an explicit entropy regularization term, where the coefficient is fixed to 0.01. Clip-higher relaxes the upper clipping bound to 0.28 while keeping the lower bound unchanged at 0.2, following the standard configuration in prior work. In addition, Clip-Cov and KL-Cov are included, using the settings reported in~\citet{cui2025entropy}.

\paragraph{Evaluation.}
We evaluate models on six widely used mathematical reasoning benchmarks: AIME24, AIME25, AMC23~\citep{numina_math_datasets}, MATH500~\citep{math500}, Minerva Math~\citep{lewkowycz2022solving}, and OlympiadBench~\citep{he2024olympiadbench}.
Following prior work~\citep{rethinking}, validation is performed with a top-p value of 0.7 and temperature 1.0 across all models
and test sets.
We report Avg@32 for AIME24 (30 problems), AIME25 (30 problems), and AMC23 (40 problems) due to their smaller size, and report Avg@1 for all other benchmarks. All evaluations are zero-shot with no additional prompts. All methods save a checkpoint every 10 steps, and the checkpoint achieving the highest average accuracy (Avg.) across benchmarks is selected for evaluation.

\begin{figure*}[t]
    \centering
    \begin{subfigure}[t]{0.32\linewidth}
        \centering
        \includegraphics[width=\linewidth]{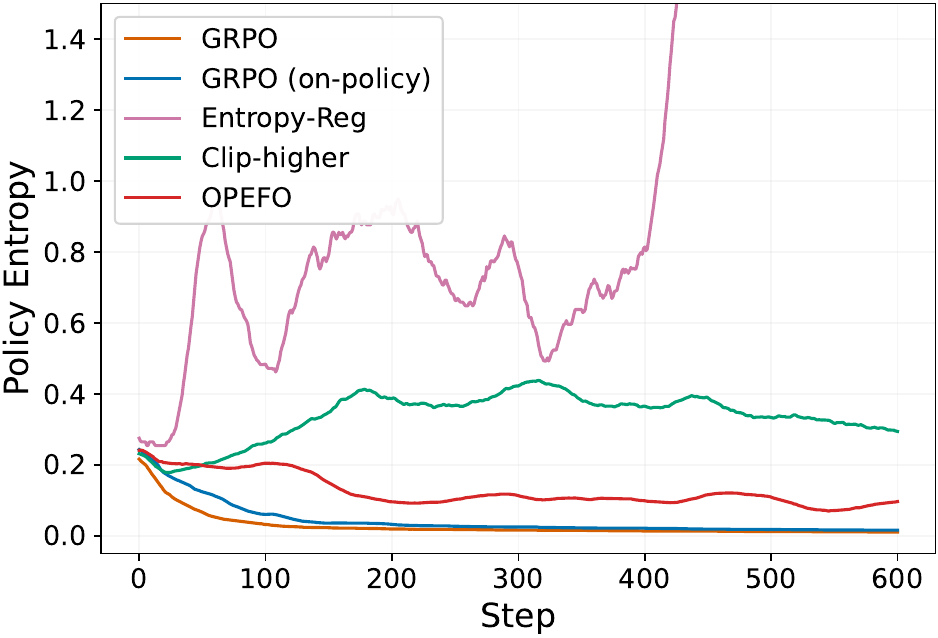}
        \caption{}
        \label{fig:main_entropy}
    \end{subfigure}%
    \hfill
    \begin{subfigure}[t]{0.32\linewidth}
        \centering
        \includegraphics[width=\linewidth]{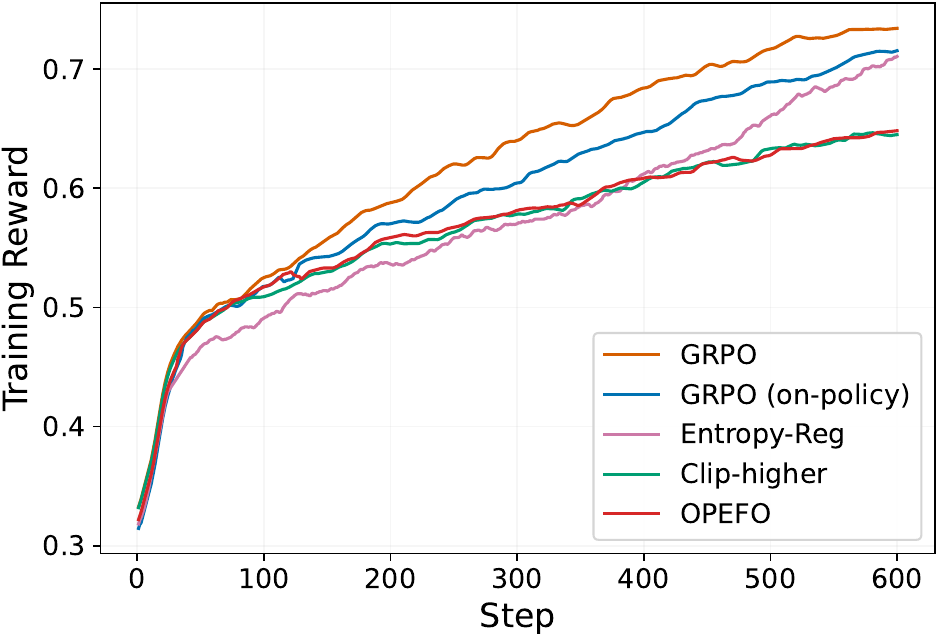}
        \caption{}
        \label{fig:main_reward}
    \end{subfigure}%
    \hfill
    \begin{subfigure}[t]{0.32\linewidth}
        \centering
        \includegraphics[width=\linewidth]{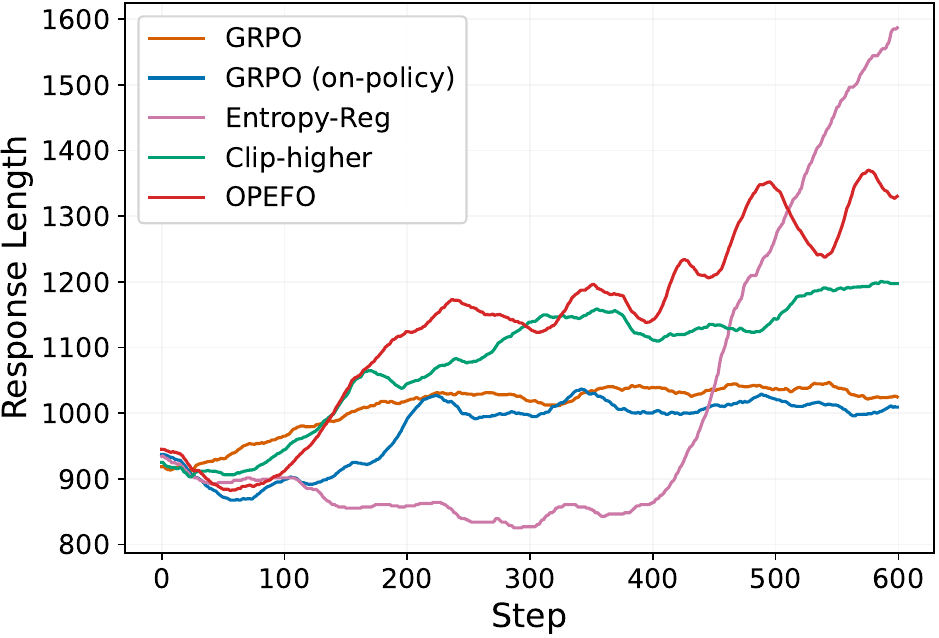}
        \caption{}
        \label{fig:main_len}
    \end{subfigure}%
    \caption{
        Training dynamics under different methods: \textbf{(a)} policy entropy, \textbf{(b)} training reward, and \textbf{(c)} response length.
    }
    \label{fig:train_dynamic}
\end{figure*}

\subsection{Main results}
Table~\ref{tab:main_res} reports the results of two base models on six mathematical reasoning benchmarks. We highlight three key observations.

First, among competitive baselines, OPEFO achieves the best overall performance. 
On Qwen2.5-Math-7B and Qwen3-Base-4B, OPEFO attains average accuracies of 52.4\% and 51.9\%, outperforming the second-best methods by 1.7\% and 1.8\%, respectively. These gains are observed across models and datasets, indicating that OPEFO improves reasoning ability in a robust manner.

Second, comparing GRPO with its strict on-policy variant reveals a clear and consistent empirical advantage of the strict on-policy setting over the approximate on-policy one. This provides empirical support for the claim discussed earlier that avoiding stale reference policies and distribution shifts can improve policy optimization in RLVR. However, strict on-policy GRPO alone, despite achieving higher accuracy, does not prevent entropy collapse, as shown in subsequent analyses.

Third, under the same strict on-policy setting, OPEFO further improves performance over strict on-policy GRPO by explicitly balancing entropy-increasing and entropy-decreasing updates, yielding average gains of 2.3\% and 1.8\% on Qwen2.5-Math-7B and Qwen3-Base-4B, respectively. This highlights the complementary role of entropy flow balancing: while strict on-policy training provides a cleaner optimization signal, OPEFO addresses the remaining instability and brings additional performance improvements under this stabilized entropy dynamics.

\begin{table*}[t]
\centering
\setlength{\tabcolsep}{3mm}
\resizebox{\textwidth}{!}{
\begin{tabular}{lcccccccc}
    \toprule
    \textbf{Method} & \textbf{AIME24} & \textbf{AIME25} & \textbf{AMC23} & \textbf{MATH500} & \textbf{Minerva} & \textbf{Olympiad} & \textbf{Avg.} \\
    \midrule
    GRPO (Strict on-policy) & 26.7 & 13.0 & 69.8 & 80.3 & 37.1 & 46.1 & 45.5 \\
    Static Scaling & 30.0 & 16.5 & 81.0 & 83.8 & 38.0 & 49.5 & 49.8 \\
    One-side ($\mathcal{S}^+$ only) & 32.8 & 18.2 & 83.1 & 84.2 & 38.7 & 47.5 & 50.8 \\
    One-side ($\mathcal{S}^-$ only) & 32.3 & 16.3 & 83.6 & 83.9 & 38.3 & 47.2 & 50.3  \\ 
    \textbf{OPEFO} & \textbf{34.5} & \textbf{19.2} & \textbf{82.2} & \textbf{85.3} & \textbf{41.6} & \textbf{51.8} & \textbf{52.4} \\
    \bottomrule
\end{tabular}}
\caption{Ablation of balancing coefficient  $\lambda^*$ on Qwen2.5-Math-7B. 
GRPO (Strict on-policy) corresponds to $\lambda^* = 0$.
``Static Scaling'' uses a fixed $\lambda^*=0.001$.
``One-side ($\mathcal{S}^+$ only)'' and ``One-side ($\mathcal{S}^-$ only)'' apply $\lambda^*$ to entropy-increasing and entropy-decreasing updates, respectively.
OPEFO applies $\lambda^*$ to both sides.}
\label{tab:ablation}
\end{table*}

\subsection{Training Dynamics Analysis}
In this section, we conduct an empirical analysis of the training dynamics of Qwen2.5-Math-7B under several representative training methods, examining how these methods affect entropy, training reward, and response length over training.

\paragraph{Overall Entropy.}
As shown on the Figure~\ref{fig:train_dynamic} (a), both standard GRPO and its strict on-policy variant exhibit a rapid collapse of policy entropy during training. While strict on-policy GRPO achieves higher accuracy than the standard setting (Table~\ref{tab:main_res}), it does not inherently prevent entropy collapse. 
In contrast, entropy regularization leads to uncontrolled entropy growth in later training stages (after approximately 400 steps), whereas Clip-higher partially mitigates collapse but still suffers from noticeable oscillations. By comparison, OPEFO maintains a smooth and stable entropy trajectory, avoiding both premature determinism and uncontrolled entropy inflation.

\paragraph{Training Reward.}
Figure~\ref{fig:train_dynamic} (b) shows steady upward trends across all methods. Among them, GRPO and its strict on-policy variant achieve the highest training reward while also exhibiting the lowest policy entropy. 
When interpreted together with the entropy curves, this pattern is consistent with premature exploitation of a limited set of high-reward answer patterns, rather than sustained exploration throughout training.
A more desirable objective is to improve training performance while maintaining sufficient entropy to preserve diversity in the learned policy. OPEFO follows this direction by stabilizing entropy while still achieving competitive reward growth.

\paragraph{Response Length.}
Figure~\ref{fig:train_dynamic} (c) shows the evolution of response length during training. For GRPO and its on-policy variant, the response length quickly saturates and stops increasing after around 200 steps, while Clip-higher stabilizes at a later stage, around 300 steps. 
Entropy regularization exhibits a sharp increase in response length after around 400 steps, coinciding with the surge in policy entropy observed in Figure~\ref{fig:train_dynamic} (a), indicating that the model begins to generate longer but increasingly unconstrained responses.
In contrast, OPEFO continues to produce longer and controlled responses throughout training. We do not claim longer responses are inherently better; rather, under the verifiable-reward setting, response length serves as a behavioral indicator of whether the model continues to explore more intermediate reasoning paths. 

\subsection{Exploration Analysis}
One of the primary motivations of controlling entropy is to improve exploration during RL training. To evaluate whether OPEFO indeed enhances exploration, we measure Pass@$k$ performance on Qwen2.5-Math-7B, which reflects the model’s ability to discover diverse reasoning paths.

We report Pass@32 on four relatively small benchmarks (Table~\ref{tab:pass32}) and additionally evaluate the scaling behavior of Pass@$k$ on AIME24 (Table~\ref{tab:passk}). As shown in Table~\ref{tab:pass32}, OPEFO consistently outperforms all baselines across datasets, indicating improved coverage of correct solutions. More importantly, Table~\ref{tab:passk} shows that OPEFO achieves higher Pass@$k$ across all $k$-values and exhibits stronger gains as $k$ increases, suggesting that it explores a broader set of valid reasoning paths rather than repeatedly generating similar responses.

\begin{table}[t]
\centering
\small
\setlength{\tabcolsep}{4pt}
\begin{tabular}{lcccc}
\toprule
Method & AIME24 & AIME25 & AMC23 & MATH500 \\
\midrule
GRPO & 59.8 & 38.4 & 93.7 & 92.5 \\
GRPO (Strict) & 61.3 & 41.2 & 94.8 & 93.3 \\
Entropy-Reg & 50.1 & 33.6 & 90.8 & 92.0 \\
Clip-higher & 57.7 & 37.6 & 94.6 & 92.7 \\
Clip-Cov & 59.0 & 41.7 & 95.3 & 93.5 \\
KL-Cov & 60.1 & 40.9 & 94.9 & 93.7 \\
\midrule
\textbf{OPEFO} & \textbf{62.4} & \textbf{43.3} & \textbf{95.6} & \textbf{94.1} \\
\bottomrule
\end{tabular}
\caption{Pass@32 performance comparison across different benchmarks.}
\label{tab:pass32}
\end{table}

\begin{table}[t]
\centering
\small
\setlength{\tabcolsep}{4pt}
\begin{tabular}{lcccc}
\toprule
Method & Pass@8 & Pass@16 & Pass@32 & Pass@64 \\
\midrule
GRPO & 45.1 & 54.3 & 59.8 & 65.6 \\
GRPO (Strict) & 50.9 & 56.2 & 61.3 & 66.5 \\
Entropy-Reg & 46.9 & 43.9 & 50.1 & 54.9 \\
Clip-higher & 50.2 & 51.7 & 57.7 & 63.3 \\
Clip-Cov & 51.3 & 53.2 & 59.0 & 64.7 \\
KL-Cov & 50.5 & 54.1 & 60.1 & 65.5 \\
\midrule
\textbf{OPEFO} & \textbf{52.5} & \textbf{56.5} & \textbf{62.4} & \textbf{68.4} \\
\bottomrule
\end{tabular}
\caption{Scaling behavior of Pass@$k$ on the AIME24 benchmark.}
\label{tab:passk}
\end{table}

\subsection{Analysis of the Balancing Coefficient $\lambda^*$}
We analyze the role of the balancing coefficient $\lambda^*$ on Qwen2.5-Math-7B, examining both different balancing strategies and the behavior of the analytically derived $\lambda^*$ during training.

\paragraph{Ablation on $\lambda^*$.}
OPEFO adaptively reweights updates by amplifying entropy-increasing updates in $\mathcal{S}^+$ and attenuating entropy-decreasing updates in $\mathcal{S}^-$ using a dynamically computed $\lambda^*$ (Eq.~\ref{eq:lambda-star}).
We first compare OPEFO against a static scaling baseline with a fixed $\lambda = 0.001$, chosen as the average value of $\lambda^*$ observed over the course of training. As shown in Table~\ref{tab:ablation}, static scaling yields moderate improvements over strict on-policy GRPO ($\lambda^*=0$), but consistently underperforms OPEFO, highlighting the importance of dynamically adjusting $\lambda^*$.

We further consider two one-sided variants that apply $\lambda^*$ only to entropy-increasing updates ($\mathcal{S}^+$ only) or only to entropy-decreasing updates ($\mathcal{S}^-$ only). Both variants outperform strict on-policy GRPO, but remain consistently inferior to full OPEFO, suggesting that jointly balancing leads to better overall performance.

\begin{figure}[t]
\centering
\includegraphics[width=0.8\linewidth]{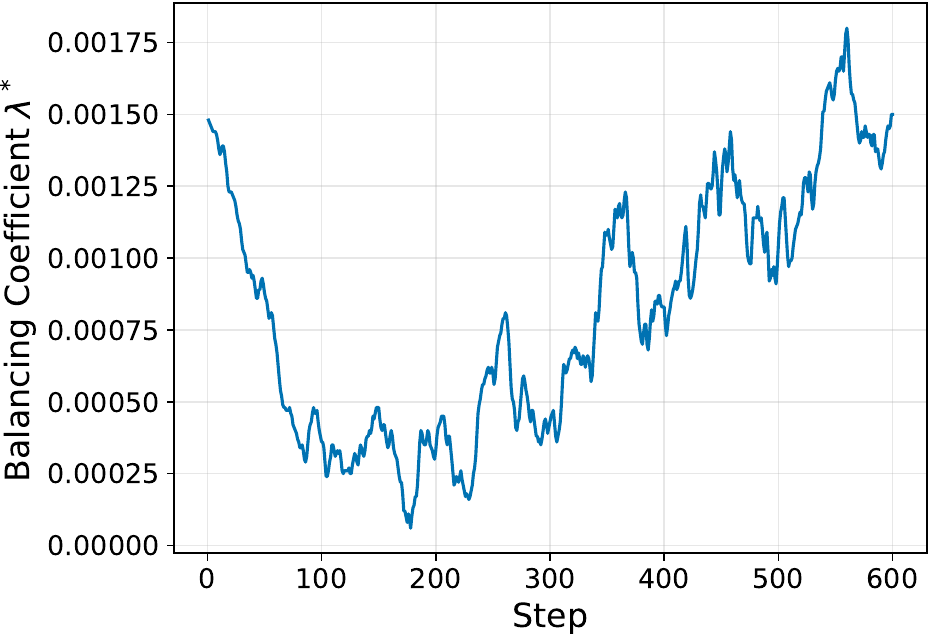}
\caption{Evolution of the balancing coefficient $\lambda^*$ over training steps.}
\label{fig:lambda}
\end{figure}

\paragraph{Dynamics of $\lambda^*$.}
Beyond ablations on balancing strategies, we further examine how the balancing coefficient $\lambda^*$ evolves during training.
As shown in Figure~\ref{fig:lambda},  $\lambda^*$ exhibits a non-stationary pattern over training. 
Overall, $\lambda^*$ remains positive throughout training, indicating a persistent need to encourage entropy-increasing updates to counteract entropy collapse.
We observe that $\lambda^*$ first decreases in the early stage and then gradually increases later in training, reflecting different entropy balancing demands across training phases.
Taken together, $\lambda^*$ displays bounded oscillations rather than converging to a fixed value, reflecting an adaptive balancing mechanism that adjusts to non-stationary entropy flow during training.

\subsection{Training Efficiency}
\begin{figure}[h]
\centering
\includegraphics[width=0.9\linewidth]{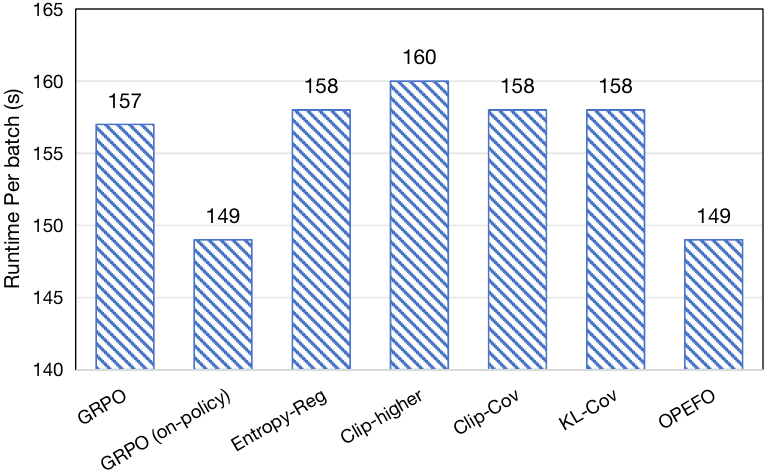}
\caption{Per-batch runtime comparison under identical rollout settings.}
\label{fig:runtime}
\end{figure}

Strict on-policy methods are often considered computationally expensive due to the need for fresh rollouts. However, in our implementation, this is not the case.
As described in Section~\ref{sec:setup}, we adopt a single update per rollout with a larger effective batch size, while approximate on-policy methods perform 8 sequential updates per rollout batch. 
This difference in update strategy directly translates into runtime. As shown in Figure~\ref{fig:runtime}, strict on-policy methods have a per-batch runtime of 149s, while approximate on-policy methods take around 158s. Although the difference is modest, it shows that strict on-policy training does not introduce additional computational overhead in practice.

\section{Conclusion}
We proposed On-Policy Entropy Flow Optimization (OPEFO), a strict on-policy mechanism for stabilizing entropy dynamics in RLVR training. By analyzing entropy collapse from a token-level entropy flow perspective, we showed that it arises from a persistent imbalance between entropy-increasing and entropy-decreasing updates. OPEFO directly addresses this issue by balancing the two opposing entropy flows without relying on reference policies or heuristic entropy regularization. Experiments on six mathematical reasoning benchmarks across two base models demonstrate that OPEFO consistently improves training stability and final performance over strong RLVR baselines.
These results suggest entropy flow balancing as a simple and effective principle for stabilizing reasoning-oriented reinforcement learning.

\section*{Limitations}
Our method offers a principled and interpretable entropy flow perspective for stabilizing RLVR training under a strict on-policy setting, and demonstrates strong empirical performance across reasoning benchmarks. We nevertheless note several limitations.

First, our analysis relies on a first-order approximation of token-level entropy change under a simplified softmax assumption, which captures dominant entropy trends but abstracts away higher-order interactions in large transformer models.

Second, the proposed entropy flow balancing mechanism operates as a local, batch-level stabilizer, where the zero entropy flow criterion serves as a sufficient condition rather than a globally optimal objective, and entropy flow behaviors may vary across tasks.

Finally, our evaluation focuses on strict on-policy RLVR for mathematical reasoning. While OPEFO is conceptually applicable to other domains and reward structures, a systematic study under dense rewards or more complex credit assignment settings is left for future work.

We view these limitations not as fundamental constraints of the proposed method, but as opportunities for extending and refining entropy-aware optimization methods in future RLVR systems.

\section*{Acknowledgments}
This research is supported by the Air Traffic Management Research Institute, NTU under the grant CAAS\_MOA\_REQ0392260\_NTU.

We only use AI-assisted tools for language polishing and improving clarity.

\bibliography{custom}

\appendix
\onecolumn
\section{Theorem Proof Details}
The following derivation is adapted from prior work~\citep{rethinking} and is included for reference and completeness.

\paragraph{Theorem 1} (First-order entropy change).
Under Assumption~1, the change
of conditional entropy between two update steps is defined as $\Delta \mathcal{H}_t \triangleq \mathcal{H}(\pi_{\theta}^{k+1}\mid s_t)-\mathcal{H}(\pi_{\theta}^{k}\mid s_t)$. Then the first-order estimation of $\Delta \mathcal{H}_t$ is:
\begin{equation}
    \Delta \mathcal{H}_{t} = 
     - \eta\, \mathbb{E}_{a \sim \pi_{\theta}^{k}(\cdot \mid s_{t})} 
    \big[\, 
    A_{t} (1 - \pi_{\theta}^{k}(a \mid s_{t}))^2 
    (\log \pi_{\theta}^{k}(a \mid s_{t}) + \mathcal{H}(\pi_{\theta}^{k} \mid s_{t})) \big]
\label{eq:app}
\end{equation}
where $\eta$ is the learning rate, and $k$ indexes the policy update step.
Note that compared to the formulation in~\citet{rethinking}, Eq.~\ref{eq:app} is specialized to the strict on-policy setting: the weight term $w_t$ (defined there as $w_t = \mathbb{I}_{\mathrm{clip}} r_t A_t$) reduces to the plain advantage $A_t$ since no importance ratio or clipping is used in our updates.

\paragraph {\textit{Proof.}} The proof is similar to that of~\citep{liu2025does}. Taking the first-order Taylor expansion, we have
\begin{equation*}
\Delta H_t  \triangleq \mathcal{H}(\pi_\theta^{k+1} \mid s_t) - \mathcal{H}(\pi_\theta^k \mid s_t)  \approx \langle \nabla_\theta \mathcal{H} (\pi_\theta^k \mid s_t), z^{k+1} - z^k \rangle
\end{equation*}
Since we have the log trick $\mathbb{E}_{a \sim \pi_\theta(\cdot \mid s)} [\nabla_\theta \log \pi_\theta(a \mid s)] = 0$, the gradient term can be derived as
\begin{equation*}
\begin{aligned}
\nabla_\theta \mathcal{H}(\pi_\theta \mid s) &= \nabla_\theta \mathcal{H}(\pi_\theta(\cdot \mid s)) \\
&= \nabla_\theta \left( - \mathbb{E}_{a \sim \pi_\theta(\cdot \mid s)} [\log \pi_\theta(a \mid s)] \right) \\
&= - \mathbb{E}_{a \sim \pi_\theta(\cdot \mid s)} [\nabla_\theta \log \pi_\theta(a \mid s) + \log \pi_\theta(a \mid s) \nabla_\theta \log \pi_\theta(a \mid s)] \\
&= - \mathbb{E}_{a \sim \pi_\theta(\cdot \mid s)} [\log \pi_\theta(a \mid s) \nabla_\theta \log \pi_\theta(a \mid s)].
\end{aligned}
\end{equation*}
Then we have
\begin{equation*}
\begin{aligned}
\Delta H_t &= \langle \nabla_\theta \mathcal{H}(\theta^k \mid s_t), (z^{k+1} - z^k) \rangle \\
&= - \left\langle \mathbb{E}_{a \sim \pi_\theta^k(\cdot \mid s_t)} [\log \pi_\theta(a \mid s_t) \nabla_\theta \log \pi_\theta(a \mid s_t)], \theta^{k+1} - \theta^k \right\rangle \\
&= - \mathbb{E}_{a \sim \pi_\theta^k(\cdot \mid s_t)} \left[ \log \pi_\theta(a \mid s_t) \langle \nabla_\theta \log \pi_\theta(a \mid s_t), \theta^{k+1} - \theta^k \rangle \right] \\
&= - \mathbb{E}_{a \sim \pi_\theta^k(\cdot \mid s_t)} \left[ \log \pi_\theta(a \mid s_t) \sum_{a' \in \mathcal{A}} \frac{\partial \log \pi_\theta(a \mid s_t)}{\partial \theta_{s_t, a'}} (\theta^{k+1}_{s_t, a'} - \theta^k_{s_t, a'}) \right] \\
&= - \mathbb{E}_{a \sim \pi_\theta^k(\cdot \mid s_t)} \left[ \log \pi_\theta(a \mid s_t) \sum_{a' \in \mathcal{A}} (\mathbf{1}\{a = a'\} - \pi(a' \mid s_t)) (\theta^{k+1}_{s_t, a'} - \theta^k_{s_t, a'}) \right] \\
&= - \mathbb{E}_{a \sim \pi_\theta^k(\cdot \mid s_t)} \left[ \left( \log \pi_\theta(a \mid s_t) - \mathbb{E}_{\tilde{a} \sim \pi_\theta^k(\cdot \mid s_t)} \log \pi_\theta(\tilde{a} \mid s_t) \right) \right. \\
& \quad \left. \left( \theta^{k+1}_{s_t, a} - \theta^k_{s_t, a} - \mathbb{E}_{a' \sim \pi_\theta^k(\cdot \mid s_t)} (\theta^{k+1}_{s_t, a'} - \theta^k_{s_t, a'}) \right) \right] \\
&= - \mathbb{E}_{a \sim \pi_\theta^k(\cdot \mid s)} \left[ [\log \pi_\theta^k(a \mid s) + \mathcal{H}(\cdot \mid s)] \left[ (1 - \pi_\theta^k(a \mid s)) (z^{k+1}_{s_t, a} - z^k_{s_t, a}) \right] \right] \\
&= - \mathbb{E}_{a \sim \pi_\theta^k(\cdot \mid s)} \left[ [\log \pi_\theta^k(a \mid s) + \mathcal{H}(\cdot \mid s)] \left[ w(s \mid a) (1 - \pi_\theta^k(a \mid s))^2 \right] \right],
\end{aligned}
\end{equation*}
where $w(s \mid a)$ is the weight in the policy gradient. \hfill $\square$

\newpage
\section{Detailed Information For Test Dataset}
\begin{table}[h]
\centering
\begin{tabular}{lcc}
\toprule
\textbf{Test Datasets} & \textbf{\#Questions} & \textbf{Level} \\
\midrule
AIME24         & 30  & Olympiad \\
AIME25         & 30  & Olympiad \\
AMC23          & 40  & Intermediate \\
MATH500        & 500 & Advanced \\
Minerva       & 272 & Graduate \\
OlympiadBench & 675 & Olympiad \\
\bottomrule
\end{tabular}
\caption{Dataset statistics.}
\label{tab:test_datasets}
\end{table}

\end{document}